\pdfoutput=1

\documentclass[11pt]{article}

\usepackage[table,dvipsnames]{xcolor}
\usepackage[]{acl}

\usepackage{times}
\usepackage{latexsym}
\usepackage{booktabs}
\usepackage{multicol}
\usepackage{tabularx}
\usepackage{graphics}
\usepackage{soul}
\usepackage[T1]{fontenc}

\usepackage[utf8]{inputenc}

\usepackage{microtype}
\usepackage{paralist}
\usepackage{diagbox}
\usepackage{multirow}
\usepackage{comment}
\usepackage{booktabs}
\usepackage{relsize}
\usepackage{array}
\usepackage{multirow}
\usepackage{graphicx}
\usepackage{xspace}
\usepackage{cleveref}
\newcommand{\ex}[1]{\textit{#1}\xspace}
\newcommand{\class}[1]{\textsf{#1}\xspace}

\newcommand{\system}[1]{\texttt{#1}\xspace}
\newcommand{\roberta}{\system{RoBERTa}}
\newcommand{\cuenb}{\system{CueNB}}
\newcommand{\z}{\phantom{0}}

\newcommand{\footnoteref}[1]{\textsuperscript{\ref{#1}}}

\title{Not another Negation Benchmark:\\ 
The NaN-NLI Test Suite for Sub-clausal Negation}

\author{Hung Thinh Truong$^{1,*}$ \qquad Yulia Otmakhova$^{1,*}$ \qquad Timothy Baldwin$^{1,3}$\\
\textbf{Trevor Cohn $^1$ \qquad Jey Han Lau$^1$ \qquad 
Karin Verspoor$^{2,1}$}\\
$^1$The University of Melbourne\,\,\, $^2$RMIT University\,\,\, $^3$MBZUAI\\
\smaller \texttt{\{hungthinht,yotmakhova\}@student.unimelb.edu.au}\,\,\,  \texttt{tb@ldwin.net} \\
\smaller \texttt{trevor.cohn@unimelb.edu.au}\,\,\,
\texttt{jeyhan.lau@gmail.com}
\,\,\,
\texttt{karin.verspoor@rmit.edu.au}}

\begin{document}

\maketitle
\begingroup\def\thefootnote{*}\footnotetext{Equal contribution}\endgroup

\begin{abstract}
Negation is poorly captured by current language models, although the extent of this problem is not widely understood.
We introduce a natural language inference (NLI) test suite to enable probing the capabilities of NLP methods, with the aim of understanding sub-clausal negation.
The test suite contains premise--hypothesis pairs where the premise contains sub-clausal negation and the hypothesis is constructed by making minimal modifications to the premise in order to reflect different possible interpretations.
Aside from adopting standard NLI labels, our test suite 
is systematically constructed under a rigorous linguistic framework.
It includes  annotation of negation types and constructions grounded in linguistic theory, as well as the operations used to construct hypotheses. This facilitates fine-grained analysis of model performance.
We conduct experiments using pre-trained language models to demonstrate that our test suite is more challenging than existing benchmarks focused on negation, and show how our annotation supports a deeper understanding of the current NLI capabilities in terms of negation and quantification.

\end{abstract}

\section{Introduction}
Negation is an important linguistic phenomenon which denotes non-existence, denial, or contradiction, and is core to language understanding.
NLP work on negation has mostly focused on detecting instances of negation \citep{peng2018negbio,khandelwal-sawant-2020-negbert, truong-etal-2022-improving}, and the effect of negation on downstream or probing tasks \citep{kassner-schutze-2020-negated, ettinger-2020-bert,hossain-etal-2020-analysis}.
A consistent finding in recent work on pre-trained language models (PLMs) is that they struggle to correctly handle negation, but also that existing NLP benchmarks are deficient in terms of their relative occurrence and variability of negation  \citep{barnes2021improving,tang-etal-2021-revisiting,hossain-etal-2022-analysis}.

In this work, we address the problem of evaluating the ability of models to handle negation in the English language using a
systematic, linguistically-based 
approach.
Specifically, we adopt the typology proposed by \citet{pullum_huddleston_huddleston_pullum_2002} whereby negation is classified based on both form (verbal and non-verbal; analytic and synthetic) and meaning (clausal and sub-clausal; ordinary and meta-linguistic).
Based on this typology, we observe that most negation instances occurring in existing benchmarks are analytic, verbal, and clausal, which is arguably more straightforward to handle than non-verbal, synthetic, and sub-clausal negation. 
For instance, the dataset proposed by \citet{hossain-etal-2020-analysis} is constructed by adding the syntactic negation cue \textit{not} to the main verb of the premise and/or the hypothesis of MNLI \citep{williams-etal-2018-broad} training examples, resulting almost exclusively in verbal, analytic, and clausal negation.

Motivated by this, we construct a new
evaluation dataset with a focus on sub-clausal negation, where it is non-trivial to determine the correct negation scope.
For instance, the negation in \ex{They saw not one but three dolphins} only scopes over the modifier \ex{one}, and thus carries a positive meaning (\ex{They saw three dolphins}).
We choose NLI as the probing task based on the intuition that a complete grasp of negation is required to make correct inference judgements.
Moreover, we adopt the test suite framework \cite{lehmann-etal-1996-tsnlp} instead of naturally-occurring text corpora, to elicit a full range of linguistic constructions that denote sub-clausal negation. This facilitates  systematic evaluation of model performance along controlled dimensions.
We collect examples for each construction from \citet{pullum_huddleston_huddleston_pullum_2002} to use as premises, and then construct corresponding hypotheses by introducing minimum changes to premises which highlight their possible interpretations.
We manually annotate the constructed pairs in terms of negation types, negation constructions, and the operations used to construct the hypotheses.

In summary, our main contributions are:
\begin{enumerate}
\item We introduce the ``NaN-NLI'' test suite for probing the capabilities of NLP models to capture sub-clausal negation.\footnote{The test suite and all code are available at \url{https://github.com/joey234/nan-nli}} In addition to standard NLI labels, it contains various linguistic annotations related to negation, to facilitate fine-grained analysis of different constructional and semantic sub-types of negation;
\item We conduct extensive experiments to confirm that our test suite is more difficult than existing negation-focused NLI benchmarks, and show how our annotations can be used to guide error analysis and interpretation of model performance; and
\item We present a subset of our test suite (NaN-Quant) with samples involving not only negation but also quantification, and show that quantification is an especially challenging phenomenon that requires future exploration.
\end{enumerate}

\section{Related Work}
To investigate the abilities of PLMs to assign the correct interpretation to negation, many probing tasks have been proposed.
For instance, \citet{kassner-schutze-2020-negated, ettinger-2020-bert} formulated a cloze-style fill-in-the-blank task where BERT is asked to predict words for two near-identical but contrasting sentences (e.g.\ \ex{A bird can \hbox to 0.5cm{\hrulefill}} vs.\ \ex{A bird cannot \hbox to 0.5cm{\hrulefill}}).
\citet{hossain-etal-2020-analysis} constructed an NLI dataset where negations essential to correctly judge the label for a premise--hypothesis pair were manually added to existing NLI benchmarks.
\citet{hartmann-etal-2021-multilingual} constructed a multilingual dataset with minimal pairs of NLI examples to analyze model behavior in the presence/absence of negation.
Most recently, \citet{hossain-etal-2022-analysis} conducted a comprehensive analysis of the effect of negation on a wide range of NLU tasks in the GLUE \citep{wang-etal-2018-glue} and SuperGLUE \citep{wang2019superglue} benchmarks.
These papers expose various limitations of both current benchmarks and PLMs in the face of negation.
However, they all focus on verbal and clausal negation, which are more straightforward to process, whereas our dataset targets non-verbal and sub-clausal negation, where it is more difficult to determine the correct scope.

The idea of using a test suite to measure the performance of NLP models was introduced by \citet{lehmann-etal-1996-tsnlp}, where the authors propose general guidelines for test suite construction. 
Adopting this methodology for a domain-specific task, \citet{cohen-etal-2010-test} constructed a dataset for benchmarking ontology concept recognition systems.
Most recently, \citet{ribeiro-etal-2020-beyond} proposed a task-agnostic testing methodology which closely follows the idea of behavioral testing from software engineering to comprehensively test the linguistic capabilities of NLP models.
The main advantages of test suites over datasets made up of naturally-occurring examples are: (1) \textit{control over the precise composition of the data}: we can undertake a targeted  evaluation of specific criteria (e.g.\ linguistic features); (2) \textit{systematicity}: a test suite has specific  structure, with samples classified into well-defined categories; and (3) \textit{control of redundancy}: we can remove samples with similar properties or over-sample rare occurrences.

\section{A Test Suite for Non-verbal Negation}

\subsection{Negation typology}
\label{sec:typology}
According to \citet{pullum_huddleston_huddleston_pullum_2002}, negation can be classified according to four main aspects: 
\begin{itemize}
    \item \textbf{Verbal vs.\ non-verbal}: verbal negation is when the negation marker is associated with the verb, while non-verbal negation is associated with an adjunct or object.
    \item \textbf{Analytic vs.\ synthetic}: when the negation marker's only syntactic function is to mark negation (e.g.\ \ex{not}), it represents analytic negation, whereas in synthetic negation the marker can have other syntactic functions (e.g.\ a compound negator \ex{nothing} can also be a subject or an object). 
    \item \textbf{Clausal vs.\ sub-clausal}: Clausal negation negates the entire clause it is contained in, whereas the scope of sub-clausal negation is strictly less than the entire clause. For instance, in \ex{Not for the first time, she felt utterly betrayed}, only the phrase \ex{Not for the first time} is negated.
    \item \textbf{Ordinary vs.\ meta-linguistic}: meta-linguistic negation acts as a correction to how the negative meaning is understood. For instance, in \ex{The house is not big, it is huge}, the negation is understood as a correction, since \ex{huge} is a more correct way of describing the size of the house. 
\end{itemize}
The first two categories relate to the syntax of negation itself while the last two relate to semantics.
In this work, we focus on sub-clausal negation as 
the correct negation scope
can be challenging to determine, which can lead to misunderstanding of the negated instance.
Although meta-linguistic negation can also cause difficulties with interpretation, as this class is rare in practice, we did not include them in our test suite. 

\subsection{Test suite construction process}

\subsubsection{Selecting premises}
\label{sec:construction-type}

We manually collect sentences from \citet{pullum_huddleston_huddleston_pullum_2002} to use as premises.
Most samples are special constructions of non-verbal negation where they denote sub-clausal negation.
Below we describe the main types of these constructions.

\textbf{\ex{Not} + quantifiers}: \ex{not} combines with a quantifier and
scopes only over that quantifier. 

\ex{Not all}: \ex{not} is used to deny the larger amount, and imply a normal value. Possible quantifiers include \ex{not all, not every, not many, not much, not often}.

\ex{Not one, not two}: \ex{not one} is used to denote a complete non-existence of something, and has the same meaning as \ex{nothing} or \ex{no one}. When combining with a numbers larger than one (usually in phrases of time and distance), \ex{not} can convey the meaning of \ex{as little as}, as in \ex{not two years ago}.

\ex{Not a little}: This construction negates the lower bound of the quantification and asserts the upper bound, denoting \ex{a fairly large amount}. For instance, \ex{not a little confusion} is equivalent to \ex{much confusion}.

\textbf{\ex{Not} + focus particles (\ex{even/only}):} 
\ex{Not even} generally marks clausal negation while \ex{not only} marks sub-clausal negation as it carries positive meaning. For instance, \ex{Not even Ed approved of the plan} implies that Ed did not approve the plan, whereas in \ex{Not only Ed approved of the plan}, Ed did in fact approve the plan.

\textbf{\ex{Not} + degree expressions}: Expressions such as \ex{not very, not quite} mark sub-clausal negation by reducing the  degree of adjectives, adverbs, or determiners (e.g.\ \ex{not very confident}).

\textbf{\ex{Not} + affixially-negated adjectives/adverbs:}
When accompanied by a gradable adjective, the construction \ex{not un-} has the meaning of negating the lower end of the scale for that adjective. For example, \ex{not unattractive} suggests the appearance ranks higher than intermediate.

\textbf{\ex{Not} in coordination:}
\ex{Not} can appear in a coordinative construction and typically scopes over only one of the coordinating parts, thus marking sub-clausal negation. In \ex{They are now leaving not on Friday but on Saturday}, \ex{not} scopes only over \ex{Friday} and denies \ex{They are leaving on Friday.}

\textbf{\ex{Not} with PPs:}
\ex{Not} can modify prepositional phrases (PPs) to denote sub-clausal negation. In \ex{Not for the first time, she felt utterly betrayed}, \ex{not} only negates the PP \ex{for the first time}, and the sentence has positive polarity in that she did feel utterly betrayed.

\textbf{\ex{Not} in verbless subordinate clauses:}
\ex{Not} can scope only over a verbless subordinate clause (e.g.\ \ex{We need someone not afraid of taking risks.}).

\textbf{\ex{Not} in implicit propositions with \ex{that}}: The construction \ex{not that} has the function of denying something that is natural or expected in the context (e.g.\ \ex{There are spare blankets in here, not that you'll have any need of them.}).

\textbf{Absolute and approximate negators:}
Absolute negators (\ex{no, never}) denote absolute non-existence but can also denote sub-clausal negation when they are part of a prepositional phrase. In \ex{They were friends in no time}, only the PP \ex{in no time} is negated. 
Approximate negators (\ex{rarely, seldom}) denote a quantification that is close to zero. 
They imply positive meaning and thus denote sub-clausal negation.

\subsubsection{Constructing premise--hypothesis pairs}
\label{construct}

When constructing hypothesis sentences for premises, we aimed to keep lexical changes to a minimum. This was especially so in the case of neutral hypotheses: though it is trivial to create any number of neutral hypotheses by changing semantically important parts of a sentence to other lexical items thus making it impossible to determine the truth value, intuitively, it would make the sentence embedding of the hypothesis quite different from that of the premise and thus easier for models to classify correctly. We also strove to make hypotheses linguistically diverse by introducing various changes to functional words rather than relying only on deletion and addition of \ex{not} as was done previously. Overall, we used 10 operations, with more than half the hypotheses including two or more changes. They are listed in \Cref{tab:changes-types} together with representative examples and their frequency counts across all sentences. 


\begin{table*}[!t]
    \centering
    \footnotesize
    \begin{tabular}{p{4cm} p{9.5cm} r}
    \toprule
        \textbf{Operation type} & \textbf{Example} & \textbf{Count} \\
    \midrule
        Indefinite quantifier change  (\ex{many, rarely}) &  \ex{She rarely goes out these days.} $\Rightarrow$ \ex{She never goes out these days.} & 74 \\
        
        Numerical quantifier change (\ex{one, twenty}) & \ex{Not for the first time, she felt utterly betrayed.} $\Rightarrow$ \ex{She felt utterly betrayed for the second time.} & 27 \\
        
        Negator addition or deletion & \ex{Not even Ed approved of the plan.} $\Rightarrow$ \ex{Even Ed approved of the plan.} & 130 \\
        
        Negator position change & \ex{He was here not ten minutes ago.} $\Rightarrow$ \ex{He was not here ten minutes ago.} & 101 \\
        
        Negator token change & \ex{Such mistakes are not common.} $\Rightarrow$ \ex{Such mistakes are uncommon.} & 6\\
        
        Clause or sub-clause deletion & \ex{Not often do we see her lose her cool like that.} $\Rightarrow$ \ex{We do not see her often.} & 36 \\
        
        Comparative quantifier change (\ex{more, less}) & \ex{They had found not one mistake.} $\Rightarrow$ \ex{They had found less than one mistake.} & 20 \\
        
        Focus particle change (\ex{even, only}) & \ex{Not even Ed approved of the plan.} $\Rightarrow$ \ex{Not only Ed approved of the plan.} & 16 \\
        
        Lexical change & \ex{We had a not very amicable discussion.} $\Rightarrow$ \ex{We did not have discussion.} & 13 \\
        
        Syntactic change & \ex{Not an accomplished dancer, he moved rather clumsily.} $\Rightarrow$ \ex{He moved rather clumsily because he was not an accomplished dancer.} & 4\\
        \bottomrule
        \end{tabular}
    \caption{Types, examples, and counts of operations used to construct hypotheses}
    \label{tab:changes-types}
\end{table*}

As outlined above, when creating hypotheses, we employed a much wider variety of linguistic operations than previous datasets, including movement of a negation marker across constituent boundaries, changing its type or scope, and substitution of indefinite pronouns. Thus we expect our dataset to be both richer and more difficult from the point of view of NLU.
On average, for each of the selected premises, we created 5 hypotheses. 


\subsubsection{Annotating the inference relationship within premise--hypothesis pairs}

Following \citet{giampiccolo-etal-2007-third}, we adopt a three-way classification of inference relationships between the premise ($p$) and the hypothesis ($q$) based on the following truth values:

\begin{itemize}
    \item {\bf\class{Entailment}}: if $p$ is True, $q$ must be True.
    
    \item {\bf\class{Contradiction}}: if $p$ is True, $q$ must be False.
    
    \item {\bf\class{Neutral}}: if $p$ is True, $q$ can be both True and False, and the available context does not allow us to make a specific judgement.
    
\end{itemize}

Two annotators (the main authors of the paper, one of whom holds a graduate degree in linguistics) labeled all constructed pairs with these categories; disagreements were resolved via discussion. 
The inter-annotator agreement prior to adjudication was 0.86 in terms of Cohen's $\kappa$ \citep{cohen1960coefficient}, which corresponds to near-perfect agreement \citep{artstein2008inter}.
We employed the following linguistic tests to distinguish between entailed and neutral pairs \citep{kroeger2018analyzing,anderson2018essentials}:
\begin{itemize}
    \item It should be impossible to deny $q$ while asserting $p$, that is, to connect $p$ and $p$ using such expressions as \ex{but it is not the fact that ...}
    
    \item It should be unnatural to express doubt about $q$ while asserting $p$, that is, to connect them using such expressions as \ex{but I am not sure whether ...}
    
    \item It should be highly redundant to assert $q$ after stating $p$, that is, to connect them with such phrases as \ex{In fact ...}
\end{itemize}

If $q$ fails at least one of these tests, it is considered to be \ex{neutral} to the premise; we regard a hypothesis to be \ex{entailed} only if it passes all three tests. A \ex{contradiction} was defined to be a statement which is the opposite of what is entailed by a premise. For example, given the premise $p = $ \ex{She didn't promise to help him}, the constructed hypotheses can be annotated in the following way:

\begin{itemize}
    \item {\bf\class{Entailment}}: \ex{She didn't promise him help} (fails all three tests).
    
    \item {\bf\class{Contradiction}}: \ex{She promised to help him} (direct opposite of $p$).
    
    \item {\bf\class{Neutral}}: \ex{She promised not to help him} (it can be be denied, asserted, and tentatively asserted).
    
\end{itemize}


\subsubsection{Annotating premise--hypothesis pairs in terms of negation types, patterns, and introduced changes}

Finally, the annotators were asked to annotate each sample with respect to the following:
\begin{itemize}
    \item \textbf{Negation types} in both the premise and hypothesis, as described in \Cref{sec:typology} (\ex{verbal} vs.\ \ex{non-verbal}, \ex{analytic} vs.\ \ex{synthetic}, \ex{clausal} vs.\ \ex{sub-clausal}).
    
    \item \textbf{Negation constructions} in the premises, as described in \Cref{sec:construction-type}. 
    For some constructions, we also specify their sub-types using their representative expressions as names. For example, for \ex{not}+quantifier, we annotate three sub-types which have distinct meanings: \ex{not many}, \ex{not one}, and \ex{not two}.
    
    \item \textbf{Operations} used to construct hypotheses, as outlined in \Cref{tab:changes-types}.
    
\end{itemize}

The initial inter-annotator agreement scores (Cohen's $\kappa$) were 0.99, 0.88, and 0.83, for negation types, negation constructions, and operations respectively, which is close to near perfect as the categories are well-defined in \citet{pullum_huddleston_huddleston_pullum_2002}.
All disagreements were then resolved via discussion.
We include such detailed linguistic annotation in the test suite to facilitate error analysis and identifying the most problematic cases. 

\subsubsection{Test suite statistics and comparison with existing negation benchmarks}

\begin{table*}[!t]
\footnotesize
    \centering
    \begin{tabular}{ p{1cm} p{1.55cm}  p{1.4cm} p{1.4cm} p{1.4cm}  p{1.4cm} p{1.4cm} p{1.4cm} p{1cm}}
    \toprule
        & & \multicolumn{3}{c}{\textbf{Premise}} & \multicolumn{4}{c}{\textbf{Hypothesis}} \\
\cmidrule(r){2-2}
\cmidrule(lr){3-5}
\cmidrule(l){6-9}
        &  \textbf{Instances} & \bf Verbal/ Non-V  & \bf Ana/Syn & \bf Clausal/ Sub-C  & \bf Verbal/ Non-V & \bf Ana/Syn & \bf Clausal/ Sub-C & \bf None  \\
    \midrule
    
    \ex{C} &  117 (45.3\%) & 5.2/ 94.9 & \z87.2/ 20.5 & 0.9/ 99.2 & 46.2/ 27.4 & 52.1/ 18.8 & 46.2/ 28.2 & 34.2\\
    \ex{E}  & \z97 (37.6\%)& 0.2/ 99.9 & \z84.5/ 20.6 & 5.2/ 94.9 & 53.6/ 20.5 & 60.8/ 11.3 & 52.6/ 21.7 & 30.9\\
    \ex{N} & \z44 (17.1\%) & 6.8/ 93.2 & 100.0/ 18.2 & 6.8/ 93.2 & 43.2/ 20.5 & 61.4/ \z2.3 & 43.2/ 20.5 & 36.4\\
    ALL & 258              & 3.5/96.5  & \z88.4/ 20.2  & 3.5/ 96.5 & 48.5/ 23.6 & 57.0/ 13.2  & 48.1/ 24.4 & 33.3\\
    \bottomrule
    \end{tabular}
    \caption{Distribution of class labels for premises-hypothesis pairs and 
 percentage of each types of negation in premises and hypotheses. \ex{C, E, N} denote \class{Contradiction}, \class{Entailment}, and \class{Neutral}, respectively.}
    \label{tab:distribution}
\end{table*}

The statistics of the resulting dataset --- named ``NaN-NLI'' --- in terms of label distribution and the types of negation used in premises and hypotheses is presented in \Cref{tab:distribution}. Following \citet{hossain-etal-2020-analysis}, we do not enforce a uniform distribution for the \class{Entailment}, \class{Contradiction}, and \class{Neutral}
classes but rather focus on constructing fluent and natural continuations which are as close to the premise as possible. 
Similarly, when constructing hypotheses, it was impossible to adhere to a particular type of negation or even to preserve it in all cases. Thus, while premises mostly have sub-clausal non-verbal negation expressed by synthetic means, the hypotheses exhibit a wider variety of patterns. It should be noted that though we report the distribution of particular negation patterns as a percentage of sentences, the values for categories do not sum to 100\% as some sentences contain more than one instance of negation.
Lastly, \Cref{tab:operation_distribution} shows the distribution of operations for each of NLI labels.
In general, we find the distribution to be quite similar for the most common categories, which allows us to claim that we are not creating major artifacts during annotation.

\begin{table}[!t]
  \footnotesize
    \centering
    \begin{tabular}{l rrr} 
    \toprule
        \textbf{Operation type} & \multicolumn{1}{c}{\textbf{C}} & \multicolumn{1}{c}{\textbf{E}} & \multicolumn{1}{c}{\textbf{N}}  \\
    \midrule
        Indefinite quantifier change & 17 & 21 & 10 \\
        Numerical quantifier change & 4 & 4 & 14 \\
        Comparative quantifier change & 4 & 4 & 8  \\
        Negator addition or deletion & 32  & 27 & 33 \\
        Negator position change & 24 & 24 & 22 \\
        Negator token change & 1 & 2 & 1 \\
        Clause or sub-clause deletion & 8 & 9 & 7 \\
        Focus particle change & 6 & 3 & 0 \\
        Lexical change & 2 & 3 & 5 \\
        Syntactic change & 0 & 2 & 0 \\
        \bottomrule
        \end{tabular}
    \caption{Distribution of operation types in each class (\%)}
    \label{tab:operation_distribution}
\end{table}

To estimate the difficulty of our benchmark relative to existing benchmarks, we use BERTScore \citep{zhang2019bertscore} to compare the average similarity between the premise and hypothesis for the three classes. For comparison, we use a subset of the MNLI dataset \citep{williams-etal-2018-broad} containing only sentences with negation, as extracted by \citet{hossain-etal-2020-analysis} (``MNLI-neg'' hereafter), and the MNLI subset of the NegNLI dataset proposed by \citet{hossain-etal-2020-analysis} (``NegNLI'' hereafter). The average similarity scores are presented in \Cref{tab:berts}; for the \class{Contradiction} and \class{Neutral} classes, in brackets we report the absolute difference over the score for the \class{Entailment} class to show how difficult it is to differentiate them. It can be seen that in our test suite, hypotheses are substantially more similar to premises than is the case for other datasets; and it is much harder to separate classes based on lexical similarity alone, with the difference between \class{Entailment} and \class{Contradiction} classes being negligible, and the difference with \class{Neutral} being smaller than for other datasets.

\begin{table}[!t]
\footnotesize
    \centering
    \begin{tabular}{p{2cm} p{1.25cm} p{1.25cm} p{1.2cm}}
    \toprule
        & MNLI-neg & NegNLI & NaN-NLI  \\
    \toprule
    
    \class{Contradiction} & 0.88 (\ex{0.02}) & 0.92 (\ex{0.00}) & \textbf{0.96} (\ex{0.00}) \\
      \class{Entailment}  & 0.91 & 0.92 & \textbf{0.96} \\
    \class{Neutral} & 0.89 (\ex{0.01}) & 0.90 (\ex{0.02}) & \textbf{0.95} (\ex{0.01}) \\
    \bottomrule
    \end{tabular}
    \caption{Average similarity (in terms of BERTScore) between premises and hypotheses for \class{Entailment}, \class{Contradiction} and \class{Neutral} classes.}
    \label{tab:berts}
\end{table}

\section{Experiments}

\subsection{Experimental settings}

For evaluation, we consider the three settings of:
\begin{itemize}
    \item \ex{Standard}: a three-way classification task with three labels: \class{Entailment}, \class{Contradiction}, and \class{Neutral}.
    \item \ex{Binary}: a binary classification task with two labels: \class{Entailment}, and \class{Not Entailment}, where we consider all \class{Contradiction} and \class{Neutral} pairs to be \class{Not Entailment}.
    \item \ex{Strict}: We only consider as correct those samples where all hypotheses for a given premise are assigned the correct label (\class{Entailment}, \class{Contradiction}, or \class{Neutral}).
\end{itemize}
We report $F_1$-score for the \ex{Standard} and \ex{Binary} settings, and {Accuracy} for the \ex{Strict} setting.
Methods investigated include \roberta \citep{liu2019roberta} and its \cuenb \citep{truong-etal-2022-improving} variant pre-trained with additional negation data augmentation and a negation cue masking strategy.
We fine-tune each model on MNLI \citep{williams-etal-2018-broad} (denoted ``-MNLI''), and the MNLI subset of the NegNLI dataset \citep{hossain-etal-2020-analysis} (denoted ``-NegNLI'').

\subsection{Main results}

For the first experiment, we measure the performance of a baseline \roberta model fine-tuned over MNLI on our test suite, in addition to other existing negation-focused NLI datasets.
As shown in \Cref{tab:benchmark_result}, the results for our evaluation set are substantially lower compared to existing NLI datasets.
This shows that our dataset contains many challenging instances of negation.
The differences are especially stark for the \class{Neutral} class, 
confirming our intuition that making the sentences in a pair as similar as possible would make them more difficult for the model.

\begin{table}[!t]
    \centering
    \footnotesize
    \begin{tabular}{p{2cm} p{1.25cm} p{1.25cm} p{1.2cm}}
    \toprule
    & MNLI-neg & NegNLI & NaN-NLI  \\
    \toprule
    \class{Contradiction} & 0.917 & 0.718 & \underline{0.664} \\
    \class{Entailment} & 0.834 & 0.656 & \underline{0.648} \\
    \class{Neutral} & 0.780 & 0.651 & \underline{0.207} \\
    All & 0.862 & 0.676 & \underline{0.580} \\
    \bottomrule
    \end{tabular}
    \caption{Results ($F_1$) of \roberta-MNLI on existing negation-focused NLI benchmarks. The lowest result for each row is \underline{underlined}.}
    \label{tab:benchmark_result}
\end{table}

\begin{figure*}[!htbp]
    \centering
    \includegraphics[width=\textwidth]{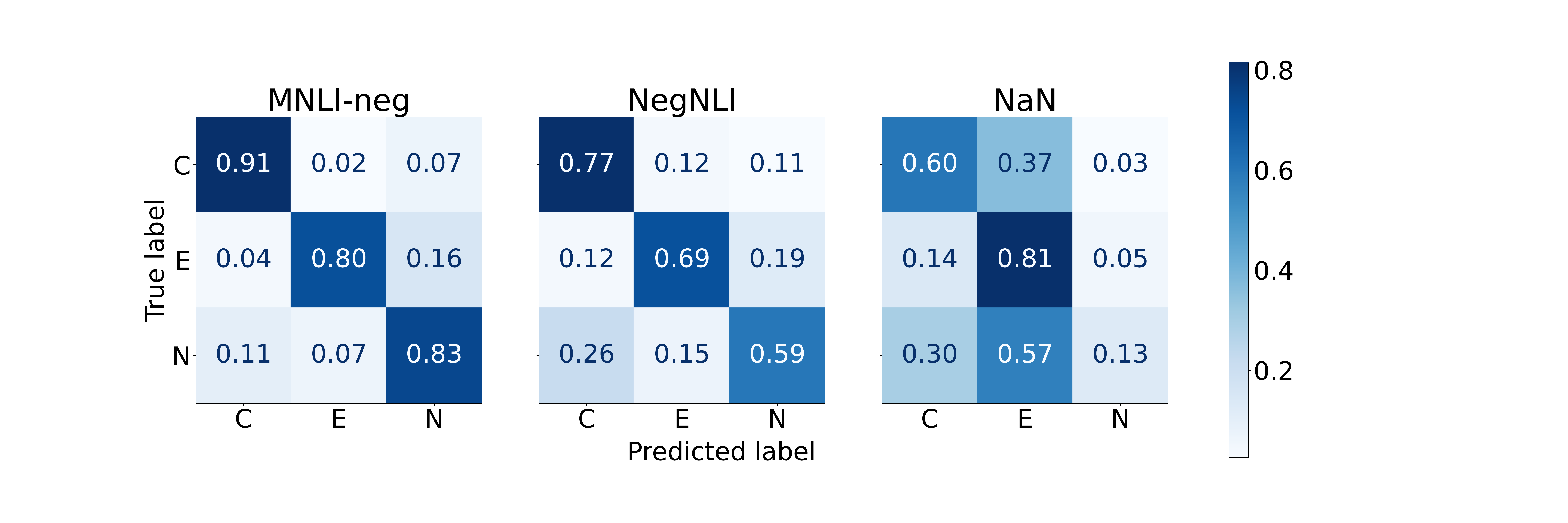}
    \caption{Confusion matrices of \roberta-MNLI on different negation-focused NLI benchmarks. \ex{C, E, N} denote the \class{Contradiction}, \class{Entailment}, and \class{Neutral} class respectively.}
    \label{fig:confusion_matrices}
\end{figure*}

\Cref{fig:confusion_matrices} provides the confusion matrices of the baseline \roberta-MNLI on existing benchmarks.
In NaN-NLI, most errors are from over-predicting \class{Entailment}.
This again shows that the sentences in our pairs are very similar lexically, and also reconfirms the known bias in MNLI that lexical overlap is a strong cue for entailment \citep{mccoy-etal-2019-right}.
On the other hand, for MNLI-neg and NegNLI, the performance for the \class{Contradiction} class is the highest.
This again reveals a bias in MNLI training data, in that if there is negation in either the premise or hypothesis, the labels are more likely to be \class{Contradiction} \citep{gururangan-etal-2018-annotation}. 

\begin{table*}[!t]
\footnotesize
    \centering
    \begin{tabular}{p{0.5cm}| p{2.5cm} p{2.5cm} p{2.5cm} p{2.5cm} p{2.5cm} }
    \toprule
    \multicolumn{2}{c}{} & \texttt{RoBERTa}-MNLI & \texttt{RoBERTa}-MNLI-NegNLI  & \texttt{CueNB}-MNLI & \texttt{CueNB}-MNLI-NegNLI  \\
    \midrule
    {\multirow{4}{*}{\rotatebox[origin=c]{90}{\textbf{\ex{Standard}}}}} & \class{Contradiction} & 0.664 & \textbf{0.692} & 0.678 & 0.651\\
    & \class{Entailment} & 0.648 & 0.684  & 0.678 & \textbf{0.694} \\
    & \class{Neutral} & 0.207 & 0.366  & 0.250 & \textbf{0.395} \\
    &  All & 0.580 & \textbf{0.629}  & 0.605 & 0.624 \\
     \midrule
    {\multirow{3}{*}{\rotatebox[origin=c]{90}{\textbf{\ex{Binary}}}}} & \class{Entailment} & 0.648 & 0.684 & 0.678 & \textbf{0.694} \\
    &  \class{Not Entailment} & 0.684 & 0.744 & 0.741 & \textbf{0.769} \\
    & All & 0.670 & 0.721 & 0.718 & \textbf{0.741} \\
    \midrule
    \multicolumn{2}{c}{\textbf{\ex{Strict}}} & 0.250 (12/48) & \textbf{0.292 (14/48)} & 0.250 (12/48) & 0.271 (13/48) \\
    \bottomrule
    \end{tabular}
    \caption{Results on our proposed NaN-NLI test suite}
    \label{tab:nan_result}
\end{table*}

\Cref{tab:nan_result} reports the detailed results for each class across different evaluation settings.
Overall, we observe a common trend in that  \cuenb outperforms the baseline \roberta when fine-tuned on the MNLI dataset.
This can be explained by the fact that \cuenb is pre-trained using more text containing negations, especially non-verbal and synthetic negations (e.g.\ \ex{no one, nobody}),
resulting in better representations for those negation cues.
Fine-tuning on the NegNLI dataset further improves performance, with both \roberta-MNLI-NegNLI and \cuenb-MNLI-NegNLI having comparable performance but \roberta performing better for the \class{Contradiction} class while \cuenb is more accurate for the \class{Neutral} class.
For the Strict setting, we observe very low results for all models with \roberta-MNLI-NegNLI outperforming its CueNB counterpart by one premise.
This
underlines the difficulty of our test suite, and shows that current methods struggle with sub-clausal negation.

\section{Discussion}
We further investigate the results of the best performing model \roberta-MNLI-NegNLI in detail to explore potential patterns in the model's predictions on our test suite.

\subsection{What types of negation are hard?}
First, we break down the results by the type of negation used in the premise or hypothesis.
There is a substantial difference in performance between samples with analytic and synthetic negation, the latter being more difficult to classify (see \Cref{sec:neg_types_results} for details). Considering that in previous datasets negation was expressed primarily by analytic means, we can conclude that the abundance of synthetic negation patterns in our dataset also contributes to its difficulty. In terms of the relation between negation types and inference labels assigned by the models, one significant\footnote{As determined by the $\chi^2$ test with $p$-value < 0.05\label{note1}} pattern we notice is that when there is no negation in the hypothesis, models assign \class{Entailment} more often.
Moreover, there is a significant\footnoteref{note1} preference to assign \class{Neutral} label when there are analytic negations in the premise compared to synthetic negation. 
We argue that this is due to the fact that \class{Neutral} is the majority class in NegNLI training data.

\begin{table*}[!t]
    \footnotesize
    \centering
    \begin{tabular}{p{5cm} p{6cm}  p{0.75cm}  p{1cm}}
    \toprule
        \textbf{Premise}  &  \textbf{Hypothesis} & \textbf{Gold} & \textbf{Predict} \\
        \midrule
        \multirow{1}{*}{Not even then did he lose patience.} & Even then, he did not lose patience. & E & E \\
         & He did not lose patience even then. & E & E \\
         & \hl{Not only then did he lose patience.} & \hl{C} & \hl{E} \\
         & \hl{Only then did he lose patience.} & \hl{C} & \hl{E} \\
         \midrule
         \multirow{1}{*}{I found his story not wholly convincing.} & I did not find his story wholly convincing. & E & E \\
         
         & \hl{I found his story wholly not convincing.} & \hl{C}  & \hl{E} \\
         & I found his story wholly convincing. & C & C \\
          &   \hl{I did not find his story wholly not convincing} & \hl{E} & \hl{C} \\
         \midrule
        
        \multirow{2}{*}{\parbox{5cm}{Not one, not two, but three of them made the mistake.}} & \hl{More than three of them made the mistake.} & \hl{C} & \hl{E} \\
        
        & More than two of them made the mistake. & E & E \\
         & More than one of them made the mistake. & E & E \\
         & \hl{One of them did not make the mistake.} & \hl{C} & \hl{E} \\
        & \hl{Two of them did not make the mistake.} & \hl{C} & \hl{N} \\
        & \hl{Less than two of them made the mistake.} & \hl{C} & \hl{E} \\
        & Less than three of them made the mistake. & C & C \\
        & Less than four of them made the mistake. & E & E \\

        \midrule
        \multirow{1}{*}{He was here not ten minutes ago.} & He was here less than ten minutes ago. & E & E \\ 
        & He was not here less than ten minutes ago. & C & C \\ 
        & \hl{He was here more than ten minutes ago.} & \hl{N} & \hl{C} \\ 
        & \hl{He was not here more than ten minutes ago.} & \hl{N} & \hl{E} \\ 
        & \hl{He was not here ten minutes ago.} & \hl{E} & \hl{C} \\ 
        & \hl{He was here one minute ago.} & \hl{C} & \hl{N} \\ 
        & He was here twenty minutes ago. & N & N \\ 
    \bottomrule
    \end{tabular}
    \caption{Selected samples along with the predictions of \roberta-MNLI-NegNLI. \hl{Highlighting} is used to  indicate prediction errors.}
    \label{tab:samples prediction}
\end{table*}

We further investigate the results based on negation constructions (\Cref{sec:construction-type}) and operations types (\Cref{construct}).
Here, we report error rate, which is the ratio of wrongly predicted samples over all samples in the same construction/modification category.
As for linguistic constructions, we find that the most difficult constructions relate to  negation in the context of a quantifier, which we further investigate in \Cref{sec:quantifier}. 
Following that, graded adjectives/adverbs, absolute and approximate negators, and degree expressions are among the more challenging  construction types for the model to handle.
On the other hand, the model deals with coordinations, implicit propositions, and verbless clauses well, with close to zero errors.
Following a similar trend, making changes to the quantifiers (either indefinite or comparative) generally confuses the model.
We find substantially high error rates for the remaining types of operation except for syntactic change, showing that the model is robust to changing the order of clauses and phrases.
\Cref{tab:samples prediction} shows some examples of P-H pairs, together with their correct and predicted labels.

\begin{table}[!t]
\footnotesize
    \centering
    \begin{tabular}{p{5cm} p{1cm}}
    \toprule
        \textbf{Construction type} & \textbf{ER} \\
    \midrule
        \ex{not} + quantifier & \textbf{0.559} \\
        \ex{not} + focus particle & 0.261\\
        \ex{not} + degree expression & 0.300 \\
        \ex{not} + affixially-negated adjective/adverb & 0.423 \\
        \ex{not} + PP & 0.067\\
        Absolute and approximate negator & 0.333 \\
        \ex{not} in verbless clause & 0.077 \\
        \ex{not} in coordination & 0.000 \\
        \ex{not} in implicit proposition & 0.000 \\
        \bottomrule
        \end{tabular}
    \caption{Error rates (ER) of negation constructions}
    \label{tab:error-construction}
\end{table}

\begin{table}[!t]
\footnotesize
    \centering
    \begin{tabular}{p{5cm} p{1cm}}
    \toprule
        \textbf{Operation type} & \textbf{ER}  \\
    \midrule
        Indefinite quantifier change & 0.486 \\
        Numerical quantifier change & 0.333 \\
        Comparative quantifier change & \textbf{0.650} \\
        Negator addition or deletion & 0.364 \\
        Negator position change & 0.327 \\
        Negator token change & 0.333 \\
        Clause or sub-clause deletion & 0.333 \\
        Focus particle change & 0.375 \\
        Lexical change & 0.308 \\
        Syntactic change & 0.000 \\
        \bottomrule
        \end{tabular}
    \caption{Error rates (ER) across operation types}
    \label{tab:error-change}
\end{table}

\subsection{Using NaN-NLI as a test suite for determining the bounds of quantification}
\label{sec:quantifier}

In over half of the samples in our test suite (133), negation interplays with quantification in terms of upper and lower bounds. In the easiest case, if a premise negates a proposition for all members of a set (\ex{None of them supported her}), a contradictory hypothesis would assert that same proposition for any number of members of the set (\ex{One of them supported her}). However, it can be hard even for humans to determine if an expression involving quantification is {True} or {False} with regard to another proposition, as it can involve not only indefinite (\ex{any, some, none, many}) and numeric (\ex{one, two, twenty}) quantifiers, but also comparative quantifiers (\ex{more, less}), gradable adjectives (\ex{attractive} $\rightarrow$ \ex{non unattractive} $\rightarrow$ \ex{not attractive} $\rightarrow$ \ex{unattractive}), or adverbs of frequency (\ex{never}, \ex{seldom}, \ex{not often}, \ex{sometimes}, etc). As negation makes this task even harder, we maintain that our test set can be a valuable resource for testing the sensitivity of models to changing of quantification bounds.

As can be seen from \Cref{tab:quant_result}, the performance of the model drops even further on the quantification subset, showing that quantification adds to the difficulty of classification. Interestingly, though, it slightly increases for the \class{Neutral} class while plummeting for the easiest class of \class{Contradiction}. We notice that often it is due to inability of the model to detect the lower or upper bound of proposition, that is, where it ceases to hold. For example, here the model correctly predicts \class{Entailment} as  \textit{more than two} is still within the quantification bounds:

\begin{quote}

    \ex{Not one, not two, but three of them made the mistake.} $\Rightarrow$ \ex{More than two of them made the mistake.}

\end{quote}

However, when we increment the number past the bound of \textit{two}, the hypothesis becomes contradictory, but the model fails to detect that and still predicts \class{Entailment}, possibly because \textit{three} is also present in the premise:

\begin{quote}

    \ex{Not one, not two, but three of them made the mistake.} $\Rightarrow$ \ex{More than three of them made the mistake.}

\end{quote}

In a similar way, such phrases as \textit{not two years ago} implicate a lower bound of the proposition, implying that it is False for numbers smaller than \textit{two}, but the model's prediction of \class{Neutral} instead of \class{Contradiction} does not reflect that:

\begin{quote}

    \ex{Not two years ago this company was ranked in the top ten.} $\Rightarrow$ \ex{One year ago this company was ranked in the top ten.}

\end{quote}

\begin{table}[!t]
\footnotesize
    \centering
    \begin{tabular}{p{2cm} p{1.5cm} p{2cm}}
    \toprule
    & NaN-NLI & NaN-Quant  \\
    \toprule
    \class{Contradiction} & 0.692 & \underline{0.477} \\
    \class{Entailment} & 0.684 & \underline{0.600} \\
    \class{Neutral} & \underline{0.366} & 0.379 \\
    All & 0.629 & \underline{0.486} \\
    \bottomrule
    \end{tabular}
    \caption{Results ($F_1$) on the whole NaN-NLI dataset vs.\ its quantification subset (NaN-Quant). The lowest result for each row is \underline{underlined}.}
    \label{tab:quant_result}
\end{table}

\subsection{Does gender affect negation?}
\label{sec:gender}
We manually augment the test suite with simple heuristics to investigate whether gender has an effect on negation.
In particular, when the sentences pairs contain a gender-specific pronouns or names, we would generate an equivalent set of sentences pairs with alternate gender pronouns or names (e.g.\ \ex{he} $\rightarrow$ \ex{she}, \ex{Ed} $\rightarrow$ \ex{Sally}).
In general, we notice no difference between the original and the gender-altered samples, showing that gender bias does not affect the types of negations in our test suite.


\subsection{Limitations}
The most prominent limitation of our test suite is unbalanced classes distribution, especially for the \class{Neutral} class. 
As discussed in \Cref{construct}, the fact that we try to construct the hypotheses by making minimum edits to the premises would make it very hard to construct meaningful \emph{Neutral} samples.
However, we argue that this is acceptable for the evaluation set, as it does not cause bias in training models.

Additionally, our test suite samples are mostly in the general English domain.
As shown in previous work \cite{khandelwal-sawant-2020-negbert,truong-etal-2022-improving}, the ways that negation is represented varies substantially across domains, and there may be other potentially challenging patterns of negation in other domains or in specific text types (e.g.\ in clinical notes), as well as other languages \citep{jimenez2021negation}.
These directions we leave for subsequent work.

\section{Conclusion}

In this work, we proposed a new test suite, dubbed NaN-NLI, for probing the performance of NLP models on data containing sub-clausal negation.
In addition to standard NLI labels, we also annotated the test suite using a systematic  linguistic framework. NaN-NLI facilitates extensive analysis of negation instances based on their negation  and construction type.
Extensive experiments show that our test suite is significantly harder for existing models than existing benchmarks, and reveal the limited capabilities of pretrained language models in dealing with this type of negation.
Detailed analysis of the results reveals a class of negations that are particularly challenging, namely those  involving quantifiers, showing that our test suite can also be used as a resource to evaluate the upper and lower bounds of quantification.

\section*{Acknowledgement}
The authors would like to thank the anonymous reviewers for their constructive reviews.
This research was undertaken using the LIEF HPC-GPGPU Facility hosted at the University of Melbourne. This Facility was established with the assistance of LIEF Grant LE170100200.
This research was conducted by the Australian Research Council Training Centre in Cognitive Computing for Medical Technologies (project number ICI70200030) and funded by the Australian Government.




\bibliography{anthology,custom}
\bibliographystyle{acl_natbib}

\appendix



\section{Implementation Details}
\label{sec:implementation}

All models are implemented using the \texttt{transformers} package from HuggingFace \cite{wolf-etal-2020-transformers}.
We use the base variant of \roberta.
For fine-tuning on NegNLI, we split the dataset into training/validation sets with a 85:15 ratio.

\begin{table}[!htbp]
    \footnotesize
    \centering
    \begin{tabular}{c c}
    \toprule
            Hyper-parameter & Value \\
        \midrule

         \texttt{batch size} & 16 \\
         \texttt{lr} & 3e-5 \\
         \texttt{epochs} & 3 \\
         \texttt{optimizer} & Adam \\
    \bottomrule
    \end{tabular}
    \caption{Hyper-parameters for fine-tuning on MNLI}
    \label{tab:hyp-mnli}
\end{table}

\begin{table}[!htbp]
    \footnotesize
    \centering
    \begin{tabular}{c c}
    \toprule
        Hyper-parameter & Value \\
    \midrule
         \texttt{batch size} & 16 \\
         \texttt{lr} & 2e-5 \\
         \texttt{epochs} & 5 \\
         \texttt{optimizer} & Adam \\
    \bottomrule
    \end{tabular}
    \caption{Hyper-parameters for fine-tuning on NegNLI}
    \label{tab:hyp-negnli}
\end{table}


\section{Results by Negation Types}
\label{sec:neg_types_results}

In \Cref{tab:neg_types_results} we show the performance of one of the models (\roberta-MNLI-NegNLI) for samples with a particular type of negation used in the premise or hypothesis. It should be noted that since in the premises negation was almost exclusively non-verbal and sub-clausal, the results for some categories (\ex{Premise - Verbal}, \ex{Premise - Clausal}) are not meaningful.

\begin{table*}[!t]
    \footnotesize
    \centering
    \begin{tabular}{p{0.5cm}| p{2.5cm}| p{2cm} p{2cm} p{2cm} }
    \toprule
     & Negation type & Precision & Recall  & $F_1$  \\
    \midrule
    {\multirow{6}{*}{\rotatebox[origin=c]{90}{\textbf{\ex{Premise}}}}} & \ex{Verbal} & 0.39  & 0.60 & 0.46 \\
    & \ex{Non-Verbal} & 0.62  & 0.59 & 0.59 \\
    & \ex{Analytic} & 0.61 & 0.59 & 0.59 \\
    & \ex{Synthetic}  & 0.43 & 0.49 & \underline{0.45} \\
    & \ex{Clausal} & 0.39 & 0.60 & 0.46 \\
    & \ex{Sub-clausal}  & 0.62 & 0.59 & 0.59 \\
     \midrule
    {\multirow{7}{*}{\rotatebox[origin=c]{90}{\textbf{\ex{Hypothesis}}}}} & \ex{Verbal} & 0.65  & 0.57 & 0.58 \\
    & \ex{Non-Verbal} & 0.63  & 0.59 & 0.57 \\
    & \ex{Analytic} & 0.68 & 0.60 & 0.60 \\
    & \ex{Synthetic}  & 0.48 & 0.45 & \underline{0.41} \\
    & \ex{Clausal} & 0.65 & 0.57 & 0.57 \\
    & \ex{Sub-clausal}  & 0.63 & 0.59 & 0.57 \\
    & \ex{None}  & 0.60 & 0.57 & 0.58 \\
    \bottomrule
    \end{tabular}
    \caption{Macro-averaged results for \roberta-MNLI-NegNLI by negation type}
    \label{tab:neg_types_results}
\end{table*}

\section{Prediction Examples}

\end{document}